\documentclass[conference]{IEEEtran}
\IEEEoverridecommandlockouts
\usepackage{cite}
\usepackage{amsmath,amssymb,amsfonts}
\usepackage{graphicx}
\usepackage{textcomp}
\usepackage{xcolor}
\def\BibTeX{{\rm B\kern-.05em{\sc i\kern-.025em b}\kern-.08em
    T\kern-.1667em\lower.7ex\hbox{E}\kern-.125emX}}
\usepackage{times}
\usepackage{latexsym}
\usepackage{amsmath, amssymb}
\usepackage{graphicx}
\usepackage{float}
\usepackage{algpseudocode} 
\usepackage{setspace}
\usepackage{caption}
\usepackage{balance}
\usepackage{etoolbox} 
\usepackage{amsmath}      
\usepackage{algpseudocode} 
\usepackage{amsfonts}     
\usepackage{graphicx}     
\usepackage{xcolor}       
\usepackage[normalem]{ulem} 
\usepackage{caption}      
\usepackage{lipsum}       
\usepackage{hyperref}     
\usepackage{fancyhdr}     
\usepackage{amsmath,amsfonts}
\usepackage[ruled,vlined]{algorithm2e}
\usepackage{amsmath,amsfonts} 
\usepackage[ruled,vlined]{algorithm2e} 
\usepackage{xcolor} 
\usepackage{multicol} 

\renewcommand{\baselinestretch}{0.95} 
\definecolor{myblue}{rgb}{0.05, 0.15, 0.6}
\definecolor{mygray}{rgb}{0.4, 0.4, 0.4}

\SetAlCapFnt{\normalfont\small\bfseries\color{myblue}}  
\SetAlCapNameFnt{\small\bfseries\color{myblue}}        
\SetKwInOut{KwIn}{\textcolor{myblue}{Input}}  
\SetKwInOut{KwOut}{\textcolor{myblue}{Output}} 
\SetKwComment{Comment}{\textcolor{mygray}{//}}{}       

\usepackage[T1]{fontenc}
\usepackage[utf8]{inputenc}

\usepackage{inconsolata}
\begin{document}

\title{KTCR: Improving Implicit Hate Detection with Knowledge Transfer driven Concept Refinement}

\author{
\IEEEauthorblockN{Samarth Garg\IEEEauthorrefmark{1}, Vivek Hruday Kavuri\IEEEauthorrefmark{2}, Gargi Shroff\IEEEauthorrefmark{2}, Rahul Mishra\IEEEauthorrefmark{3}}
\IEEEauthorblockA{\IEEEauthorrefmark{1}ABV-IIITM Gwalior, India\\
Email: imt\_2020085@iiitm.ac.in}
\IEEEauthorblockA{\IEEEauthorrefmark{2}IIIT Hyderabad, India\\
Email: \{kavuri.hruday, gargi.shroff\}@research.iiit.ac.in}
\IEEEauthorblockA{\IEEEauthorrefmark{2}IIIT Hyderabad, India\\
Email: rahul.mishra@iiit.ac.in}
}


\maketitle

\begin{abstract}
The constant shifts in social and political contexts, driven by emerging social movements and political events, lead to new forms of hate content and previously unrecognized hate patterns that machine learning models may not have captured. Some recent literature proposes data augmentation-based techniques to enrich existing hate datasets by incorporating samples that reveal new implicit hate patterns. This approach aims to improve the model's performance on out-of-domain implicit hate instances. It is observed, that further addition of more samples for augmentation results in the decrease of the performance of the model. In this work, we propose a Knowledge Transfer-driven Concept Refinement method that distills and refines the concepts related to implicit hate samples through novel prototype alignment and concept losses, alongside data augmentation based on concept activation vectors. Experiments with several publicly available datasets show that incorporating additional implicit samples reflecting new hate patterns through concept refinement enhances the model's performance, surpassing baseline results while maintaining cross-dataset generalization capabilities.\footnote{DISCLAIMER: This paper contains explicit statements that are potentially offensive.}
\end{abstract}


\section{Introduction}
Addressing and curbing hateful and abusive content is essential in today's web environment. On social media platforms like X (twitter.com) and forums such as Reddit (reddit.com), an unprecedented volume of posts containing hate speech is generated daily \cite{waseem-2016-racist}. Dealing with this content manually is infeasible, making it necessary for these platforms to implement automated hate detection models \cite{schmidt-wiegand-2017-survey,qian-etal-2018-hierarchical, Founta, vidgen-etal-2021-learning, yang-etal-2024-uncertainty} to mitigate harmful content effectively. However, the constantly evolving social and political landscape, driven by new movements and events, gives rise to novel forms of hate content and reveals previously unrecognized hate patterns that machine learning models may not have captured \cite{nejadgholi2022improving}. As a result, these models need to be continuously retrained with newly collected datasets to keep up with emerging types of hate content.

\begin{figure}[t]
  \centering
  \includegraphics[width=0.27\textwidth]{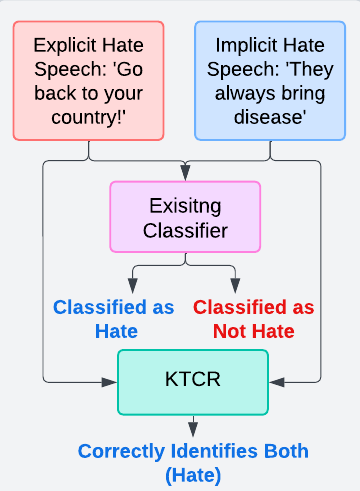} 
\caption{\small{Comparison of the performance of traditional classifiers and the proposed KTCR method. Traditional classifiers often correctly identify explicit hate speech (e.g., "Go back to your country!") but fail to detect implicit hate speech (e.g., "They always bring disease"). The KTCR method, however, successfully identifies both explicit and implicit hate speech as hate, improving overall detection accuracy.}}
\vspace{-0.5cm}
  \label{fig:intro_diagram}
\end{figure}

To this end, some recent efforts have been made in the literature \cite{nejadgholi2022improving} by utilizing data augmentation techniques to enrich the original dataset with samples having novel hate speech patterns. Augmentation is performed using Testing Concept Activation Vector (TCAV) \cite{kim2018interpretability} based sensitization towards human-defined concepts such as implicit hate patterns. Implicit hate patterns are more subtle, indirect, and hard to capture by the model \cite{waseem-etal-2017-understanding, nejadgholi2022improving}. However, when the size of the augmentation data surpasses a certain threshold, the model's performance starts to decline, even on the original test set, resulting in a loss of the advantages gained through implicit sample augmentation. 

To tackle this challenge, we introduce the Knowledge Transfer-driven Concept Refinement method (KTCR), which distills and refines concepts associated with implicit hate samples through innovative prototype alignment and concept loss strategies inspired by a recent computer vision work \cite{gupta2024concept}, along with data augmentation based on Concept Activation Vectors (CAVs). While previous studies have examined data augmentation and domain adaptation to enhance implicit hate speech detection, our approach uniquely combines knowledge transfer with concept refinement. Utilizing a teacher-student framework, KTCR improves the model's sensitivity to implicit hate speech by effectively transferring nuanced patterns from the teacher model to the student model.

In contrast to earlier methods that mainly focus on augmenting data with implicit examples or correcting biases, our approach actively distills and refines concepts related to implicit hate. This enables the student model to internalize subtle patterns of implicit abuse more effectively. Additionally, our method ensures robust performance in explicit hate speech detection, addressing the challenge of integrating new information without compromising existing capabilities.

The key contributions of this work are: 1) To the best of our knowledge, we are the first to introduce concept refinement based on Concept Activation Vectors (CAVs) for hate speech detection tasks.
2) We perform experiments using multiple benchmark datasets to assess the proposed technique's effectiveness for cross-dataset generalization.
3) We conduct several ablations of the model to evaluate the contribution of each component.
4) We perform a rigorous error analysis and examine the impact of concept refinement at the gradient level.

\section{Methodology}

\begin{figure*}[h!]
  \centering
  \includegraphics[width=\textwidth]{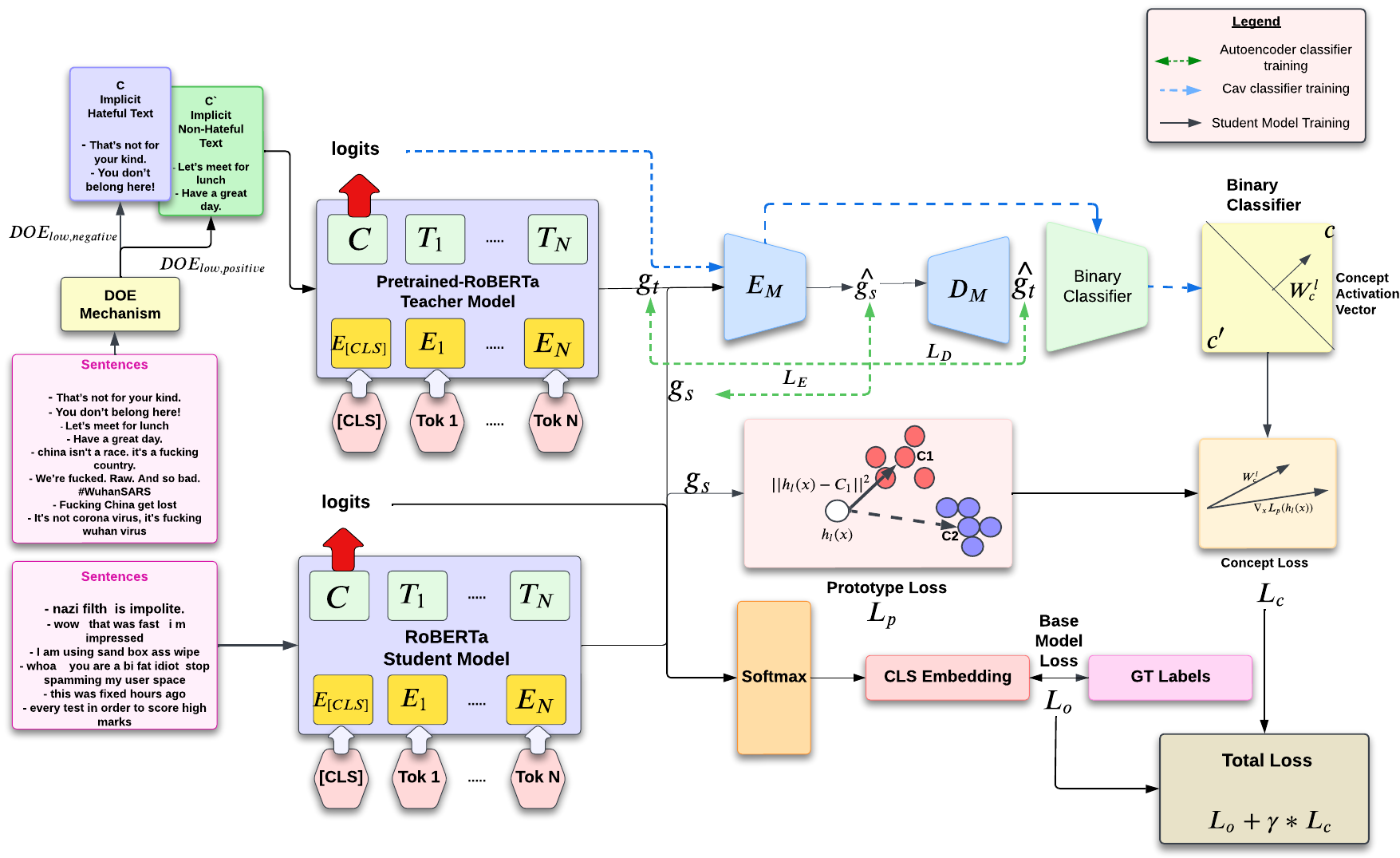} 
  \caption{Overview of the Knowledge Transfer via Concept Refinement (KTCR) framework. The teacher model ($\mathbf{M_T}$), trained on explicit hate samples, guides the student model ($\mathbf{M_S}$) to refine its understanding of implicit hate through autoencoder-based activation mapping, prototype alignment, and concept loss. The autoencoder maps the teacher's activations ($g_t$) to the student's activations ($g_s$), while prototype alignment is used to refine the learned representations.
}
  \label{fig:concept_refinement}
\end{figure*}

We begin by curating two distinct datasets: a concept set ($\mathcal{C}$) and a random set ($\mathcal{R}$). The concept set includes sentences that subtly convey hate, while the random set contains non-hateful sentences. Both sets are derived from the EA dataset, with implicitly hateful or non-hateful sentences identified using the Degree Of Explicitness (DOE) method \cite{nejadgholi2022improving}.

\subsection{Degree Of Explicitness (DOE)}

The DOE \cite{nejadgholi2022improving}, an extension of the TCAV method \cite{kim2018interpretability}, quantifies the model's sensitivity to explicitness by measuring how a new example impacts the model's perception of explicitness. A high sensitivity implies explicitness, while reduced sensitivity indicates implicitness. By calculating the DOE for each potential example, we prioritize sentences with the lowest DOE scores (i.e., the most implicit) for data augmentation. These implicit sentences are then split into two groups—hateful ($C$) and non-hateful ($C^*$) based on the class labels—to create the concept set $\mathcal{C}$ and random set $\mathcal{R}$, respectively. This approach enables the model to learn from more nuanced, implicitly hateful content. The algorithm for the DoE is provided in the \hyperref[sec: appendix]{appendix}.
\begin{algorithm}[ht]
\small 
\caption{\textbf{Knowledge Transfer via Concept Refinement (KTCR)}}
\label{alg:ktcr}
\KwIn{\textit{Frozen Teacher Model} $\mathbf{M_T}$, \textit{Student Model} $\mathbf{M_S}$, \textit{Autoencoder} $\mathbf{A_E}$, \textit{Datasets} $\mathcal{C}$, $\mathcal{R}$, $\mathcal{W}$ (Wiki)}
\KwOut{\textit{Trained student model} $\mathbf{M_S}$ for implicit hate detection}
\BlankLine

\textbf{\underline{Step 1: Data Preparation}}  \\
\textit{Input:} $\mathcal{C}$ (Concept Set), $\mathcal{R}$ (Random Set), and $\mathcal{W}$ (Wiki Dataset) \\
Load $\mathcal{C}$ and $\mathcal{R}$ for concept training and initialize $\mathcal{W}$ as the input for $\mathbf{M_S}$.

\textbf{\underline{Step 2: Initialize Models}}  \\
Load pre-trained \textbf{teacher model} $\mathbf{M_T}$ (frozen) and \textbf{student model} $\mathbf{M_S}$. Initialize \textbf{classification model} $O$ and \textbf{Autoencoder} $\mathbf{A_E}$ to map teacher activations $\mathbf{g_t}$ to student activations $\mathbf{g_s}$ as follows:

$A_E: E_M(g_t) \rightarrow D_M(g_s)$
\textbf{\underline{Step 3: Concept Distillation Loop}}\\
\For{\textit{each epoch} $\mathbf{N}$ \textit{from} 1 \textbf{to} $\mathbf{N_{\text{max}}}$}{
    \textbf{Step 3.1: Autoencoder and CAV Training}  \\
    \If{\textit{CAV update cycle is reached}}{
        Train autoencoder $\mathbf{A_E}$ using the losses: 

        $L_D = \|g_t - \hat{g_t}\|^2 \quad \text{and} \quad L_E = \|\hat{g_s} - g_s\|^2$
        Update the \textbf{classification model} where CAVs $\mathbf{w^l_C}$ are extracted from the learned concept embeddings.
    }   
    \textbf{Step 3.2: Prototype Alignment}  \\
    \If{\textit{Prototype update cycle is reached}}{
        For each sample $x$, group activations $h_l(x)$ into $K$ clusters. Calculate the \textbf{Prototype Alignment Loss}: 
        
        $L_p(x) = \frac{1}{K} \sum_{k=1}^{K} \|h_l(x) - C_k\|^2$
        Update prototypes $C_k$ using the update rule:
        
        $C_{n+1}^k = (1 - \beta)C_n^k + \beta C_c^k$
    }
    \textbf{Step 3.3: Train Student Model}\\
    Train $\mathbf{M_S}$ on Wiki dataset $\mathcal{W}$ using the \textbf{combined loss} function:
        $L = L_o + \gamma L_C $
where 
    
    $L_C(x) = \left| \cos \left( \nabla L_p(h_l(x)), w_C^l \right) \right|$
}
\textbf{\underline{Step 4: Final Output}} \\
\textit{Return the trained student model} $\mathbf{M_S}$ with improved detection of implicit hate speech.
\end{algorithm}

\subsection{Knowledge Transfer via Concept Refinement (KTCR)}

To enhance the effectiveness of our classifier in recognizing implicit abusive content, we introduce the Knowledge Transfer via Concept Refinement (KTCR), which builds upon the concept distillation method initially proposed for computer vision use-cases \cite{gupta2024concept} for desensitization of the model for a particular concept. KTCR is crucial for transferring human-centered concepts, such as implicit and explicit abusive content, into the model, improving its interpretability and enabling better generalization to new data. Traditional training approaches \cite{mathew2021hatexplain, hoang2024toxcl, plaza2023respectful, wang2024cltl}, tend to optimize performance on specific datasets but often struggle to internalize nuanced, abstract concepts crucial for real-world tasks, such as detecting implicit hate speech.

To address this challenge, KTCR involves transferring knowledge from a teacher to a student model. The teacher model is trained to distinguish between concepts and random examples, guiding the student model in aligning its internal representations with the target concepts defined by the teacher. Through this refinement process, the student model becomes more proficient at identifying subtle forms of abuse, enhancing its overall ability to detect implicit hate speech.

\subsection{Autoencoder for Activation Mapping}

We utilize the [CLS] embeddings from the teacher model to guide the learning process of the student model through a pre-trained autoencoder. The autoencoder addresses a key challenge, which is the discrepancy between the activation spaces of the teacher and student models.

The autoencoder is composed of an encoder ($E_M$) and a decoder ($D_M$), and its purpose is to map the teacher model’s activations into a space where the student model can process them effectively. By bridging the gap between the teacher's and student’s activation spaces, the student model is better able to learn how to distinguish between hateful and non-hateful content, including nuanced and implicit forms of hate speech.

The learning objective for the autoencoder involves minimizing a combined loss function that includes $L_{D}$ and $L_{E}$. Here, $L_{D}$ refers to the L2 loss between the original teacher activations ($g_t$) and their reconstructed versions ($\hat{g}t$), which ensures the accurate reconstruction of the teacher's learned representations. $L_{E}$, on the other hand, refers to the L2 loss between the mapped teacher activations ($\hat{g}_s$) and the actual student activations ($g_s$). By minimizing these losses across a concept set of text samples, the student model effectively transfers and internalizes the teacher model’s understanding of both explicit and implicit abusive content.

\subsection{Prototype Alignment Loss}

The student model (RoBERTa base) is trained using samples from the Wiki dataset, and the training objective consists of two main components: the \textit{concept loss} and the \textit{base model loss}. We use categorical cross-entropy for the base model loss, which focuses on classifying input text as either hateful or non-hateful. To enhance concept learning in the intermediate layers (which is where concepts tend to be more effectively learned than in the final classification head), we use class prototypes as pseudo-labels, building on the approach introduced in \cite{snell2017prototypical}.

For each sample $x$, the activation of layer $l$ is denoted as $h_l(x)$. We group the activations for each class into $K$ clusters, with the centroid of each cluster, $C_k$, acting as the class prototype. The Prototype Alignment Loss, $L_p(x)$, is then computed as the squared distance between a sample’s activation and its respective class prototype:

\[
L_p(x) = \frac{1}{K} \sum_{k=1}^{K} \|h_l(x) - C_k\|^2.
\]

This prototype loss quantifies the extent to which a sample's activation is aligned with its class prototype, capturing both sample-specific and class-wide patterns. The prototypes are periodically updated to reflect changes in the activation space. The update rule for the prototype centroids is as follows:

\[
C_{n+1}^k = (1 - \beta)C_n^k + \beta C_c^k,
\]

where $\beta$ controls the rate at which the prototype centroids are updated.

\subsection{Concept Loss}

The concept loss serves the purpose of fine-tuning a neural network model by adjusting its sensitivity to specific human-centered concepts. For example, if a model places too much emphasis on an irrelevant feature (e.g., common profanities like "damn" or "hell," which are not necessarily abusive in context), the concept loss can steer the model away from this feature by desensitizing it to the irrelevant concept.

The concept loss $L_C(x)$ quantifies the alignment between the loss gradient and the Concept Activation Vector (CAV). This is computed as the cosine of the angle between the gradient of the prototype loss and the CAV direction, denoted as $w^l_C$:

\[
L_C(x) = |\cos(\nabla L_p(h_l(x)), w^l_C)|.
\]

When the gradient aligns with the CAV direction, $L_C(x)$ reaches its minimum, making the model more sensitive to the concept of interest. In the context of implicit hate detection, minimizing the loss term $1 - L_C(x)$ increases the model’s sensitivity to implicit hate.

\subsection{Combined Loss Function}

To further fine-tune the model, the student model is trained over multiple epochs using a combined loss function:

\[
L = L_o + \gamma L_C,
\]

where $L_o$ represents the base loss (e.g., cross-entropy), and $\gamma$ is a parameter that controls the balance between the concept loss $L_C$ and the base loss. This combined loss function optimizes the model's ability to detect implicit abusive content while maintaining strong performance on the base task of identifying explicit and non-hateful content.As outlined in Algorithm~\ref{alg:ktcr}, the KTCR framework facilitates the transfer of knowledge from the teacher model to the student model through a series of refinement steps.

\section{Experiments}
We demonstrate the effect of our Knowledge Transfer-driven Concept Refinement on the generalization of models, specifically in response to evolving datasets driven by real-world social and political events. In this section, we present the experiments conducted to evaluate the proposed approach. We focus on how it compares to existing state-of-the-art methods and perform a series of ablation studies. These studies help us understand the contribution of each component to the overall results. The datasets used, experimental setup, evaluation metrics, and insights from our ablation studies are detailed below.

\subsection{Datasets Used}
We utilize three datasets in our experiments: the \textbf{Wikipedia Toxicity} \cite{wulczyn2017ex} (Wiki) dataset for pre-COVID hate speech, and the \textbf{East-Asian Prejudice} \cite{DBLP:journals/corr/abs-2005-03909} (E.A) and \textbf{Covid-Hate} \cite{he2021racism} (C.H) datasets for post-COVID hate speech. For binary classification, we binarize each dataset as follows: the "Toxic" class in the Wiki dataset, "Hostility against an East Asian (EA) entity" class in the E.A dataset, and "Stigmatizing" class in the C.H dataset are treated as \textbf{positive} labels. The "Normal" class in the Wiki dataset, all the other classes in the E.A dataset, and the "Not stigmatizing" class in the C.H dataset are considered \textbf{negative} labels.

\subsection{Experimental Setup}

We employ a \textbf{RoBERTa-Large} model as the teacher model ($\mathbf{M_T}$), trained on a balanced, augmented set of implicitly hateful samples from the \textbf{E.A dataset} using the Degree of Explicitness (DoE) methodology. The student model ($\mathbf{M_S}$) is a \textbf{RoBERTa-Base} model pretrained on the \textbf{Wiki dataset} for general toxic content and fine-tuned on balanced implicitly hateful samples from the E.A dataset, serving as our concept set ($\mathcal{C}$) in the KTCR framework. An autoencoder ($\mathbf{A_E}$) maps activations from the teacher to the student model, using a learning rate of 0.005 and training for 3 epochs. The student model is trained with a learning rate of $3 \times 10^{-5}$, a concept loss weight $\gamma=5$, and for 3 epochs. We use k-means clustering ($k=2$) for prototype alignment. Experiments are conducted on a 48GB Nvidia RTX 6000 Ada GPU with batch size 32, taking approximately 20 minutes after dataset balancing. Further experimental results with varying augmentation sizes and class balances are discussed in Section~\ref{sec: result}.

\subsection{Evaluation Metrics}
To evaluate our binary classification model, we use the \textbf{F1-score (macro)} and \textbf{Area Under the ROC Curve (AUC)}. The \textbf{F1-score (macro)} provides a balanced measure of precision and recall across both classes, accounting for potential class imbalances. The \textbf{AUC} evaluates the model's ability to distinguish between classes, with higher values indicating better discrimination. Together, these metrics offer a comprehensive view of both the classifier's overall performance and its class separation ability.

\section{Ablation Studies}

To evaluate the contribution of each component in our proposed \textbf{Knowledge Transfer via Concept Refinement (KTCR)} framework, we conducted a series of ablation studies to understand how different training strategies and data configurations affect the model's performance in detecting implicit hate speech.

\textbf{Effect of Concept Refinement Without Student Exposure to Implicit Hate.} In the first experiment, we assessed the impact of applying Concept Refinement when the student model had no prior exposure to implicit hate examples during training. Specifically, we trained the teacher model ($\mathbf{M_T}$) on 3,000 implicitly hateful samples from the EA dataset, selected based on low DoE scores. The student model ($\mathbf{M_S}$) was trained solely on the Wiki dataset, which mainly contains explicit hate speech and non-hateful content. The same 3,000 implicitly hateful samples were used as the concept set ($\mathcal{C}$) for Concept Refinement. Applying KTCR increased the F1-score on the EA test set from 0.27 (baseline) to 0.61, but the AUC remained around 0.68. This indicates that without prior exposure to implicit hate, the student model's ability to discriminate between classes is limited, even with Concept Refinement.

\textbf{Introducing Implicit Hate Examples to the Student Model.} To assess the importance of the student model's exposure to implicit hate, we augmented its training data with 3,000 implicitly hateful samples from the EA dataset (the same used for the teacher model and concept set), balancing the dataset. Applying KTCR with these balanced datasets increased the F1-score on the EA test set to 0.72 and the AUC to 0.81. This demonstrates that prior exposure to implicit hate significantly enhances the student model's detection capabilities, further improved by Concept Refinement.

\textbf{Impact of Augmentation Size.} We investigated the effect of increasing the augmentation size to 5,000 and 8,000 samples from the EA dataset, with imbalanced classes. As shown in Table~\ref{tab:ablation_aug_size}, although the F1-score remained at 0.72, the AUC improved with larger augmentation sizes, reaching 0.81 with 8,000 samples. This suggests that larger, augmentation sets enhance the model's discrimination between classes.

\begin{table}[H]
\centering
\scalebox{0.72}{
\begin{tabular}{lcc}
\hline
\textbf{Augmentation Size} & \textbf{F1-score} & \textbf{AUC} \\ \hline
3,000 samples              & 0.72              & 0.76        \\
5,000 samples              & 0.72              & 0.78         \\
8,000 samples              & 0.72              & 0.81         \\ \hline
\end{tabular}
}
\caption{Performance on the EA test set with varying augmentation sizes after applying KTCR.}
\label{tab:ablation_aug_size}
\end{table}

\textbf{Effect of Class Imbalance.} To test the impact of class imbalance, we re-balanced the augmented datasets to have equal numbers of hateful and non-hateful samples and re-applied KTCR. Balancing increased the AUC from 0.81 to 0.91, indicating better class discrimination, while the F1-score slightly decreased from 0.72 to 0.69. Moreover, convergence time gets reduced from approximately 4 hours to 20 minutes. This confirms that class imbalance negatively affects performance and efficiency.

\textbf{Importance of Concept Refinement.} To evaluate the importance of Concept Refinement, we trained the student model on the augmented datasets without KTCR. This model achieved an F1-score of 0.61 and an AUC of 0.81. With KTCR applied, performance improved to an F1-score of 0.72 in one of the configurations and an AUC of 0.91 in another, highlighting the significant contribution of Concept Refinement. The ablation studies demonstrate that prior exposure to implicit hate examples is crucial for the student model to effectively detect implicit hate speech. Increasing the augmentation size improves discrimination ability, as reflected in higher AUC scores. Balancing the datasets enhances model performance and significantly reduces convergence time. Overall, the KTCR framework is essential for enhancing the student model's capability to detect implicit hate speech. These experiments highlight the importance of integrating Concept Refinement with appropriate data preparation strategies to achieve optimal results in implicit hate speech detection.

\subsubsection{Analysis of Activation and Gradient Norms}

\begin{figure}[h!]
    \centering
    \includegraphics[width=0.50\textwidth]{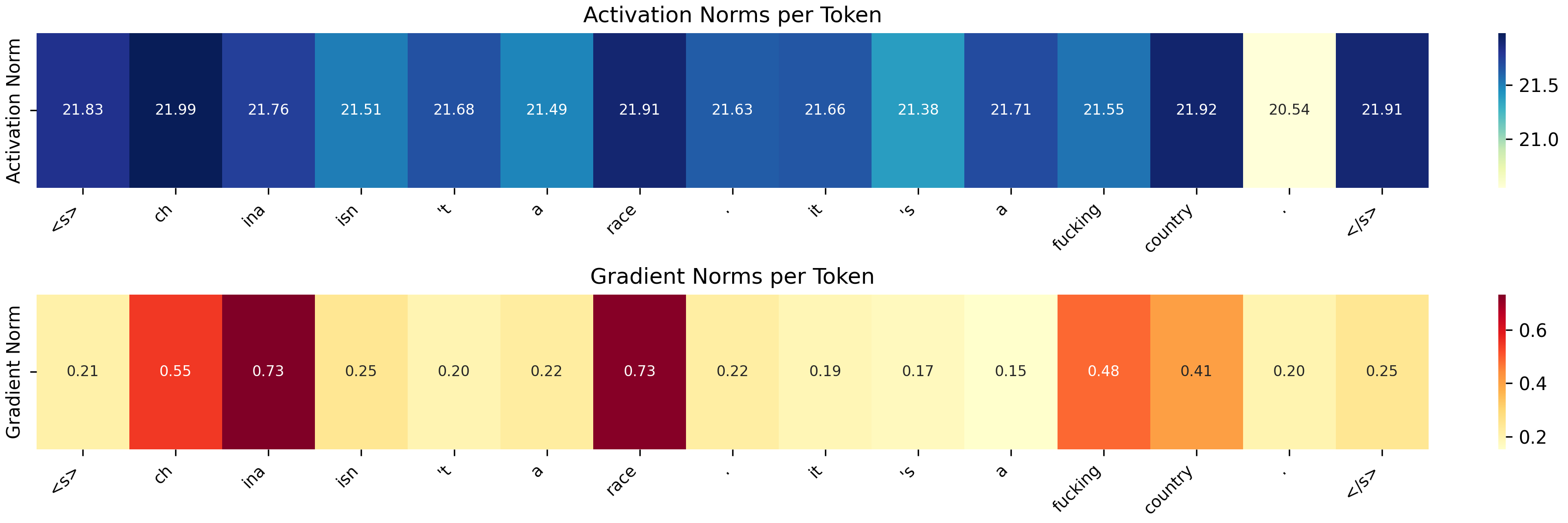}
    \caption{Activation and Gradient Norms Before Applying KTCR}
    \label{fig:activation_norms}
\end{figure}

\begin{figure}[h!]
    \centering
    \includegraphics[width=0.50\textwidth]{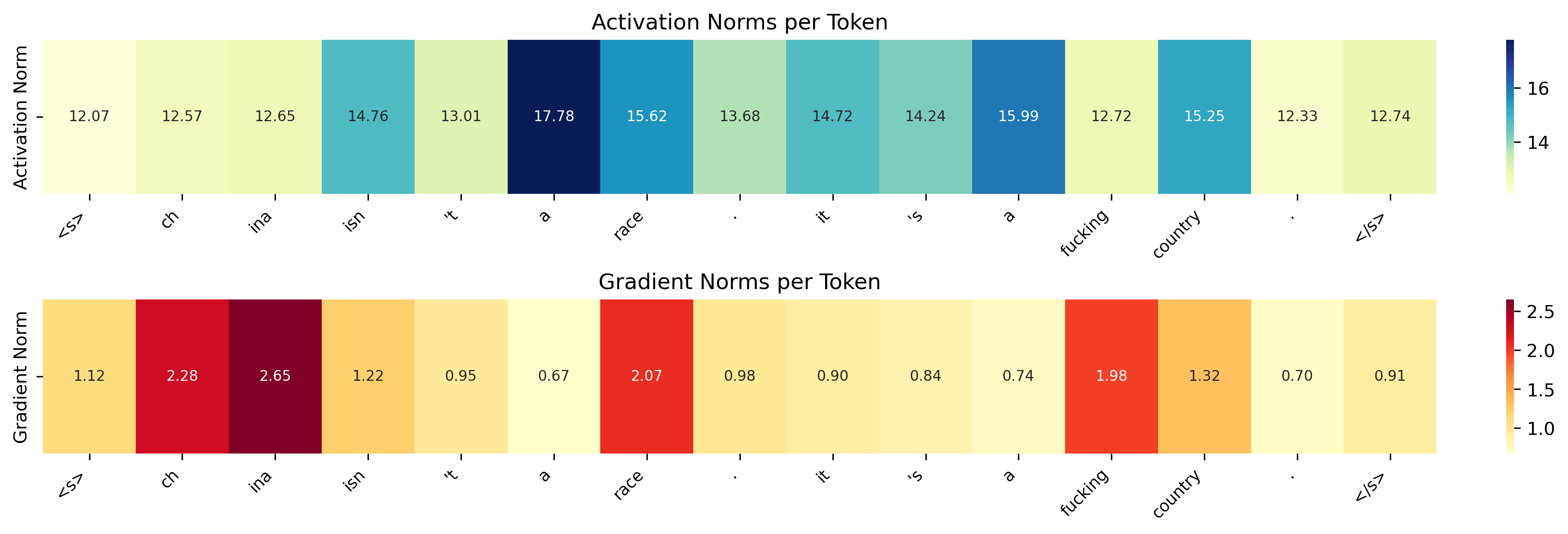}
    \caption{Activation and Gradient Norms After Applying KTCR}
    \label{fig:gradient_norms}
\end{figure}

To assess the impact of KTCR on the model's internal representations, we analyze the activation and gradient norms of the last layer of RoBERTa before and after applying KTCR. Figures~\ref{fig:activation_norms} and~\ref{fig:gradient_norms} display these norms for tokens in a sample input text. Before applying KTCR (Figure~\ref{fig:activation_norms}), activation norms are relatively uniform across tokens, ranging from 20.54 to 21.99. Tokens like ``<s>'', ``ch'', and ``ina'' have norms around 21.5--21.9, while ``country'' is slightly lower at 20.54.

After applying KTCR (Figure~\ref{fig:gradient_norms}), activation norms exhibit greater variability, ranging from 12.07 to 17.78. Tokens such as ``race'' and ``a'' show higher norms (17.78 and 17.5), indicating increased importance, while ``<s>'', ``ch'', and ``ina'' have lower norms (12.07 to 12.65), reflecting a shift in focus.

Gradient norms also change significantly. Initially, they range from 0.15 to 0.73, with ``race'' having the highest (0.73). After KTCR, gradient norms increase substantially, ranging from 0.67 to 2.65. Tokens like ``ina'', ``ch'', and ``race'' have higher values (2.65, 2.28, and 2.07), indicating larger parameter updates during training.

\paragraph{Interpretation}

The increased variability in activation and gradient norms suggests that KTCR adjusts the model to assign greater importance to tokens associated with implicit hate speech. This refinement enhances the model's sensitivity to nuanced, context-dependent abusive content, improving the detection of implicit hate speech. 
\subsection{Error Analysis}

To evaluate the effectiveness of our proposed KTCR method in implicit hate speech detection, we conducted an error analysis on test samples from the EA dataset where our model corrected misclassifications made by the baseline model. Specifically, we focused on 112 instances where the KTCR model successfully detected hate speech that the baseline model missed.

\textbf{Detection of Implicit Hate Speech:} The KTCR model showed significant improvement in identifying implicit hate speech. For instance, in Example 1 (Table~\ref{tab_errors}), the hashtags like \#ChinaVirus and \#WuhanCoronavirus implicitly link the virus to a specific nationality, which may fuel xenophobia. The baseline model, likely focusing on explicit indicators of hate, failed to recognize this bias, whereas the KTCR model correctly identified the implicit sentiment.

\textbf{Recognition of Coded Language:} Several corrected instances involved coded language or slurs that the baseline model misclassified. In Example 2, the term \#kungflu is a derogatory play on words mocking Chinese culture. The KTCR model's enhanced conceptual representation allowed it to capture such coded hate speech more effectively.

\textbf{Handling of Sarcasm and Irony:} Some tweets employed sarcasm or irony to convey hate, which the baseline model struggled to detect. In Example 3, the sarcastic reference to China as the "great growth driver" alongside hashtags like \#wuhanflu and \#chinavirus implied blame and ridicule, which the KTCR model accurately identified as hate speech. The KTCR method improves the detection of subtle, implicit hate speech through its enhanced conceptual understanding, allowing it to identify nuanced patterns such as sarcasm, irony, and coded language, which the baseline model often missed.

\begin{table}[H] 
\centering 
\small 
\begin{tabular}{p{0.95\linewidth}} 
\hline 
\textbf{Example 1}: \textit{At least 30 million people on lockdown after virus death toll rises.... \#ChinaVirus \#CoronaVirus \#WuhanCoronavirus \#China \#Wuhan \#WHO \#CoronoaVirus} \\ 
\hline 
\textbf{Example 2}: \textit{Leave it to Americans to meme a world war and a virus that has 50 million people quarantined in China... \#kungflu} \\ \hline
\textbf{Example 3}: \textit{The great growth driver of the world... the birthplace of \#wuhanflu \#chinavirus} \\
\hline
\end{tabular}
\caption{Examples where the baseline model misclassified implicit hate speech, but the KTCR model corrected it.} 
\label{tab_errors}
\end{table}

\section{Results}
\label{sec: result}
Table \ref{table:results1}, \ref{results: Table2} and \ref{results: Table3} presents the F1-scores and AUC of classifiers updated using Concept Refinement, DoE, and confidence-based augmentation methods on EA, Wiki and CH datasets respectively across different augmentation sizes. Our method consistently outperforms DoE and other baselines in terms of F1 Score and AUC for different augementation sizes, demonstrating that Concept Refinement effectively adapts the model to new data while maintaining its performance on previous data. \\
\begin{table}[H]
\centering
\scalebox{0.72}{
\begin{tabular}{lccll}
        & \multicolumn{2}{c}{Aug. set} &  &  \\ 
        \hline
        {Method}  & Teacher & Student & F1-score & AUC \\
        \hline
        Concept Refinement & 5K EA & 3K EA & 0.69 & \textbf{0.91}\\
         & 3K EA & 3K EA & 0.61 & 0.84 \\
         & 3K EA & - & 0.61 & 0.68 \\
         & 5K EA & - & 0.61 & 0.67 \\
         & 8K EA & 3K EA & \textbf{0.72} & 0.81 \\
 \hline \hline
        DoE      & - & 3K EA & 0.61 & 0.81 \\
        Conf.    & - & 4K EA & 0.54 &  0.69 \\
        Merging  & -    & EA & 0.58 & 0.72 \\
        Baseline & -     & - & 0.27 & 0.64 \\
         \hline
    \end{tabular}
 }   
    \caption{AUC and F1-scores for Concept Refinement Method and various augmentation methods, as well as the original Wiki classifier as a baseline on EA dataset.}
    \label{table:results1}
\end{table}

\textbf{Effectiveness of Concept Refinement for New Abuse Types:} For the EA dataset, which contains implicit abusive speech of new type, Concept Refinement demonstrates superior performance across all augmentation sizes. This is evident from the higher F1 scores and AUC values across the board when compared to other methods. Concept Refinement achieves an F1-score of 0.69 and an AUC of 0.91 when using a 5K EA + 3K EA augmented set, which suggests that Concept Refinement is highly effective at adapting models to new data types (such as new types of abuse). \\

\begin{table}[H]
\centering
\scalebox{0.72}{
\begin{tabular}{lccll}
        & \multicolumn{2}{c}{Aug. set} &  &  \\ 
        \hline
        {Method}  & Teacher & Student & F1-score & AUC \\
        \hline
        Concept Refinement & 5K EA & 3K EA & 0.76 & 0.93\\
         & 3K EA & 3K EA & 0.77 & 0.93 \\
         & 3K EA & - & 0.89 & 0.96 \\
         & 5K EA & - & \textbf{0.89} & \textbf{0.97} \\
          \hline \hline
        DoE      & - & 3K EA & 0.82 & 0.96 \\
        Conf.    & - & 4K EA & 0.71 &  0.94 \\
        Merging  & -    & EA & 0.78 & 0.94 \\
        Baseline & -     & - & 0.82 & 0.96 \\
         \hline
    \end{tabular}
  }  
    \caption{AUC and F1-scores for Concept Refinement Method and various augmentation methods, as well as the original Wiki classifier as a baseline on the Wiki dataset.}
    \label{results: Table2}
\end{table}

\textbf{Maintaining model performance on the Wiki dataset:} Concept Refinement also maintains strong performance when applied to the original Wiki dataset, displaying results comparable to DoE. The best-performing model using the 5K EA augmentation set achieves an AUC of 0.97, surpassing other methods and baselines. Additionally, other augmentation sets yield comparable performance, further highlighting that Concept Refinement can handle the challenge of learning new information while retaining the model’s performance on older, established data. \\

\begin{table}[H]

\centering
\scalebox{0.69}{
\begin{tabular}{lcccccc}
        & \multicolumn{2}{c}{Aug. set} & \multicolumn{2}{c}{F1-score} & \multicolumn{2}{c}{AUC} \\ 
        \hline
        Method & Teacher & Student & CH & CH* & CH & CH* \\
        \hline
        Concept Refinement & 3K EA & 3K EA & - & 0.42 & - & 0.55 \\
        & 5K EA & 3K EA & - & 0.46 & - & \textbf{0.65} \\
        & 3K EA & - & - & \textbf{0.54} & - & 0.58 \\
        & 5K EA & - & - & 0.53 & - & 0.58 \\
         \hline \hline
        DoE & - & 3K EA & \textbf{0.73} & - & \textbf{0.78} & - \\
        Conf. & - & 4K EA & 0.71 & - & 0.75 & - \\
        Merging & - & EA & 0.72 & - & 0.75 & - \\
        Baseline & - & - & 0.69 & - & 0.74 & - \\
         \hline
    \end{tabular}
 }   
    \caption{AUC and F1-scores for Concept Refinement Method and various augmentation methods, as well as the original Wiki classifier as a baseline on CH and CH* dataset.}
    \label{results: Table3}
\end{table}

** The performance of the models on the CH dataset cannot be directly compared due to inconsistencies in the dataset. The CH dataset referenced \cite{nejadgholi2022improving} is unavailable, and the version found online contains approximately 5,000 tweets, while it was claimed to consist of 2,300 tweets.

\section{Related Work}

\subsection{Implicit Hate Speech Detection}

Detecting hate speech is a well-established task in natural language processing, but most existing methods focus on explicit expressions of hate \cite{schmidt2017survey, fortuna2018survey}. Implicit hate speech, which conveys hateful content in subtle, indirect, or context-dependent ways, poses a significant challenge due to its reliance on shared cultural knowledge and context \cite{waseem2017understanding, caselli2020feel}. Traditional lexicon-based approaches often fail to capture implicit abuse, as highlighted by \cite{wiegand2019detection}, because such methods depend on overt hateful language.

Several studies have attempted to address the detection of implicit hate speech. \cite{breitfeller2019finding} found that inter-annotator agreement is lower for implicit abuse due to its subtlety and proposed improved annotation practices to capture nuanced content. \cite{wiegand2021implicitly} created a dedicated dataset for implicit abuse and conducted a contrastive analysis using various linguistic features to enhance detection methods. These works emphasize the need for advanced techniques that go beyond surface-level text analysis.

\subsection{Generalization and Domain Adaptation in Hate Speech Detection}

Generalizability is a critical issue in hate speech detection systems, as models often struggle to perform well across different datasets or in the face of evolving language \cite{swamy2019studying, fortuna2020toxic}. Cross-dataset evaluations by \cite{swamy2019studying} and topic modeling approaches by \cite{nejadgholi-kiritchenko-2020-cross} revealed significant drops in performance when models are tested on unseen data, indicating over-fitting to specific datasets. Data augmentation has been explored to enhance the robustness of hate speech classifiers. \cite{badjatiya2019stereotypical} proposed adding benign examples containing biased terms to mitigate model biases toward specific identity terms. \cite{karan2018cross} utilized multi-task learning and domain adaptation techniques to improve generalization across different types of abusive content. \cite{field2020unsupervised} introduced propensity matching and adversarial learning to encourage models to focus on implicit bias indicators rather than surface-level cues. These methods aim to reduce the model's reliance on spurious correlations present in the training data.

\subsection{Concept-Based Interpretability and Knowledge Transfer}

Concept Activation Vectors (CAVs), introduced by \cite{kim2018interpretability} in the context of computer vision, have been adapted for interpretability in NLP tasks \cite{ghorbani2019towards}. CAVs enable the quantification of a model's sensitivity to human-understandable concepts, providing insights into the model's decision-making process.

In hate speech detection, \cite{nejadgholi2022improving} leveraged the concept of the \textit{Degree of Explicitness} (DoE) to guide data augmentation. By incorporating samples with low DoE scores—representing implicit hate—they improved the model's ability to generalize to new, unseen types of hate speech. Their approach highlights the importance of carefully curated datasets that include a diverse range of hate expressions.

Building on TCAV \cite{kim2018interpretability}, \cite{gupta2024concept} introduced a concept desensitization approach for digit recognition in the computer vision domain. Their method distills knowledge from a teacher model to a student model using a concept loss, effectively reducing the model's dependence on spurious features like digit color. In contrast, we sensitize the student model to novel, previously unseen 
 latent concepts, specifically, implicit hate patterns in post-COVID hate speech. This represents a novel application of such methods in the NLP domain. Furthermore, we refine knowledge using concept loss for a pretrained language model (RoBERTa), as opposed to a simple multilayer perceptron (MLP).

\section{Conclusion}
Keeping up with the constantly evolving nature of hate speech content is crucial for machine learning models to remain effective, necessitating periodic re-training and ongoing dataset curation. In this work, we introduce a Knowledge Transfer-driven Concept Refinement (KTCR) method that distills and refines concepts related to implicit hate, without relying on large curated datasets, while also preserving the model's original capabilities. Rigorous experiments with publicly available datasets and ablation studies demonstrate that the proposed model outperforms state-of-the-art methods and shows strong cross-dataset generalization. In the future, we plan to benchmark the KTCR model for other relevant downstream tasks, such as misinformation and clickbait detection.
\section{Limitations}

The proposed Knowledge Transfer-driven Concept Refinement (KTCR) method, while effective for implicit hate detection in English, it is not benchmarked in multilingual and low-resource language settings, which may pose a unique set of challenges. 
Furthermore, the model has not been tested with code-mixed content, often seen on social media, where multiple languages are blended within a single post. To ensure a fair comparison, we use only the RoBERTa base model, as this is consistent with the state-of-the-art approach. We did not explore larger language models such as the Llama and GPT series, so the impact of kTCR on these models remains uncertain.

\section*{Ethical Considerations}

This research utilizes publicly available datasets solely for non-commercial academic purposes. All data have been anonymized to protect individual privacy, ensuring that no personal identities are disclosed. We have taken care to mitigate potential biases in the datasets and the developed model, striving for fairness and equity in hate speech detection. Additionally, we acknowledge the dual-use nature of hate speech detection technologies and emphasize that our work is intended to support responsible moderation efforts, not to infringe on free speech. By adhering to ethical guidelines and prioritizing privacy and fairness, we aim to conduct our research responsibly and ethically.

\bibliographystyle{IEEEtran}
\bibliography{references}

\vspace{12pt}
\section*{Appendix} \label{sec:appendix}

\begin{algorithm}[ht]
\small 
\caption{\textbf{Degree of Explicitness (DOE) Calculation for Balanced Dataset}}
\KwIn{EA dataset (sentences with labels), number of sentences $k$}
\KwOut{Concept set (implicitly hateful), Random set (implicitly non-hateful)}

\textbf{Initialize:} \\
\hspace{0.5cm} Concept set = $\{\}$, Random set = $\{\}$ \\
\hspace{0.5cm} DOE\_list\_hateful = $[\ ]$ \\
\hspace{0.5cm} DOE\_list\_nonhateful = $[\ ]$ \\

\For{each sentence $x$ in EA dataset}{
    \textbf{Calculate DOE score using TCAV-based methodology:} \\
    \hspace{0.5cm} (i) Define explicit class examples as baseline. \\
    \hspace{0.5cm} (ii) Evaluate impact of sentence $x$ on model's sensitivity to explicitness. \\
    \hspace{0.5cm} (iii) Compute DOE score $DOE(x)$ for $x$. \\

    \If{$x$ is labeled as \textbf{hateful}}{
        Append $(x,\ DOE(x))$ to DOE\_list\_hateful.
    }
    \ElseIf{$x$ is labeled as \textbf{non-hateful}}{
        Append $(x,\ DOE(x))$ to DOE\_list\_nonhateful.
    }
}

\textbf{Sort} DOE\_list\_hateful in ascending order based on DOE scores.

\textbf{Sort} DOE\_list\_nonhateful in ascending order based on DOE scores.

\textbf{Select} top $k/2$ sentences from DOE\_list\_hateful and add them to \textbf{Concept set} (implicitly hateful).

\textbf{Select} top $k/2$ sentences from DOE\_list\_nonhateful and add them to \textbf{Random set} (implicitly non-hateful).

\Return{Concept set and Random set}

\end{algorithm}

\end{document}